\documentclass{article}

\usepackage{microtype}
\usepackage{graphicx}
\graphicspath{ {./figures/} }
\usepackage{subcaption}
\usepackage{booktabs} %
\usepackage{todonotes}
\usepackage[frozencache=true,cachedir=.]{minted}
\usepackage{listings}
\usepackage{svg} 
\usepackage{pdfpages}
\usepackage[T1]{fontenc}  %

\usepackage{hyperref}

\newcommand\code[1]{\texttt{#1}}
\newcommand{\ChunkStore}{\code{ChunkStore}}
\newcommand{\RateLimiter}{\code{RateLimiter}}
\newcommand{\RateLimiters}{\code{RateLimiter}s}
\newcommand{\Selector}{\code{Selector}}
\newcommand{\Selectors}{\code{Selector}s}
\newcommand{\Sampler}{\code{Sampler}}

\newcommand{\Remover}{\code{Remover}}
\newcommand{\Table}{\code{Table}}
\newcommand{\Tables}{\code{Table}s}
\newcommand{\Item}{\code{Item}}
\newcommand{\Items}{\code{Item}s}
\newcommand{\Chunk}{\code{Chunk}}
\newcommand{\Chunks}{\code{Chunk}s}
\newcommand{\Writer}{\code{Writer}}
\newcommand{\SPI}{SPI}
\newcommand{\MinSize}{\code{min\_size}}
\newcommand{\ReverbDataset}{\code{ReverbDataset}}
\newcommand{\NumSampledItems}{\code{num\_sampled\_items}}
\newcommand{\NumInsertedItems}{\code{num\_inserted\_items}}

\newcommand{\TFData}{\code{tf.data}}

\newcommand{\BeginPython}{
\begin{minipage}[]{\linewidth}
\begin{minted}[fontsize=\footnotesize]{python}
}
\newcommand{\EndPython}{
\end{minipage}
\end{minted}
}

\usepackage[final]{mlsys2020}

\mlsystitlerunning{Reverb: A Framework for Experience Replay}

\begin{document}

\twocolumn[
\mlsystitle{Reverb: A Framework for Experience Replay}

\mlsyssetsymbol{equal}{*}

\begin{mlsysauthorlist}
\mlsysauthor{Albin Cassirer}{dm}
\mlsysauthor{Gabriel Barth-Maron}{dm}
\mlsysauthor{Eugene Brevdo}{gbus}
\mlsysauthor{Sabela Ramos}{gbzhr}
\mlsysauthor{Toby Boyd}{gbus}
\mlsysauthor{Thibault Sottiaux}{dm}
\mlsysauthor{Manuel Kroiss}{dm}
\end{mlsysauthorlist}

\mlsysaffiliation{dm}{DeepMind, London, UK}
\mlsysaffiliation{gbus}{Google Research, Mountain View, California, USA}
\mlsysaffiliation{gbzhr}{Google Research, Zurich, Switzerland}

\mlsyscorrespondingauthor{Albin Cassirer}{cassirer@google.com}

\mlsyskeywords{Machine Learning}

\vskip 0.3in

\begin{abstract}
A central component of training in Reinforcement Learning (RL) is Experience: the data used for training. 
The mechanisms used to generate and consume this data have an important effect on the performance of RL algorithms.

In this paper, we introduce Reverb: an efficient, extensible, and easy to use system designed specifically for experience replay in RL. 
Reverb is designed to work efficiently in distributed configurations with up to thousands of concurrent clients.

The flexible API provides users with the tools to easily and accurately configure the replay buffer.
It includes strategies for selecting and removing elements from the buffer, as well as options for controlling the ratio between sampled and inserted elements.
This paper presents the core design of Reverb, gives examples of how it can be applied, and provides empirical results of Reverb's performance characteristics.

\end{abstract}
]

\printAffiliationsAndNotice{}  %

\section{Introduction}
\label{introduction}
Experience plays a key role in Reinforcement Learning (RL); how best to use this data is one of the central problems of the field.
As RL agents have advanced in recent years, taking on bigger and more complex challenges
such as Atari, Go, StarCraft, and Dota
\citep{mnih2015human, silver2016mastering, vinyals2019grandmaster, openai2019dota},
the generated data has grown in both size and throughput.
To accelerate data collection, many RL experiments split the agent into
two distinct parts:
data generators (\textit{actors}) and data consumers (\textit{learners}),
which are run in parallel \cite{horgan2018distributed,hoffman2020acme}.
However, data must now be stored and transported between generators and consumers.
Efficiently transporting and storing this data is in itself a challenging engineering problem.

To address this challenge, we present Reverb: an efficient, extensible, and easy
to use system for experience transport and storage.
Reverb is designed to be flexible, making it suitable to be used as Experience Replay (ER) \cite{lin1992self} or Prioritized Experience Replay (PER) \cite{schaul2015prioritized}, which is a crucial component in a number of off-policy algorithms including Deep Q-Networks \cite{mnih2015human}, Deep Deterministic Policy Gradients \cite{lillicrap2015continuous}, and Soft Actor-Critic \cite{haarnoja2018soft}.
Reverb is equally suitable as a FIFO, LIFO, or Heap-based queues, enabling on-policy methods like Proximal Policy Optimization \cite{schulman2017proximal} and IMPALA \cite{espeholt2018impala}.

Another strength is its efficiency, making it well suited for large-scale
RL agents with many actors and learners running in parallel. Researchers
have used Reverb to run experiments with thousands of concurrent actors and learners~\cite{menger}.
This scalability, coupled with Reverb’s flexibility, allows researchers to investigate problems that require different scales without worrying about changing infrastructure components.

Reverb provides an easy-to-use mechanism for controlling the ratio of sampled to inserted data elements. This ratio is commonly used to set the number of gradient updates that are taken per data element, which can significantly impact RL algorithm performance \cite{fedus2020revisiting, fu2019diagnosing,hoffman2020acme}.
Controlling this feature is easy in simple synchronous settings. However, it is challenging to implement when clients are running concurrently in a distributed system.
With Reverb, users can control the relative rate of data collection to training in RL experiments regardless of scale.

The remainder of the paper is organized as follows. Section~\ref{sec:related_work}
discusses related work and alternatives to Reverb. Section~\ref{sec:design} describes Reverb's core components and how they contribute to building an efficient data storage and transportation system for RL. Code examples are provided in
Section~\ref{sec:examples}. Section~\ref{sec:performance} is a discussion of
Reverb's performance and empirical benchmark results. Finally, in
Section~\ref{sec:conclusion}, we provide concluding remarks.

More information, source code, and examples are available in Reverb’s GitHub repository\footnote{\url{http://github.com/deepmind/reverb}}.
For RL algorithm implementations using Reverb, the agents
in Acme\footnote{\url{http://github.com/deepmind/acme}}~\cite{hoffman2020acme}
and TF-Agents\footnote{\url{http://github.com/tensorflow/agents}}~\cite{TFAgents},
are excellent references.

\section{Related Work}
\label{sec:related_work}
Many existing RL libraries and frameworks provide implementations of (Prioritized) Experience Replay; including RLlib \cite{liang2018rllib}, garage \cite{garage}, stable-baselines \cite{stable-baselines}, and Tianshou \cite{tianshou}.
These implementations enable researchers to run off-policy reinforcement learning experiments.

SEED RL \cite{Espeholt2020SEED}, a scalable reinforcement learning agent, provides performant
implementations of ER and PER for off-policy RL and a FIFO queue for on-policy RL. 
However, these implementations are designed for the SEED architecture.
In SEED, only the learner inserts and samples from the replay buffer
so the samples per insert ratio (described in Section~\ref{sec:rate_limiting})
is strictly managed by the learner. Besides,
data is only sharded when using multiple learners, and there is no
possibility of having multiple tables referencing the same data.

We created Reverb to address the scaling or flexibility requirements of our research teams that the
aforementioned options did not meet. On one hand, they needed to scale to a number of clients and QPS that 
were not possible without sharding across multiple machines and handling concurrent
actors (experience generators) and learners (consumers). On the other hand, our researchers also required the
ability to switch between ER or PER for off-policy and FIFO queues for on-policy data without substantial
code changes, as well as having the ability to control the ration between samples and insertion.

\begin{figure}[t]
\centering
    \begin{subfigure}{\columnwidth}
        \includegraphics[scale=0.6]{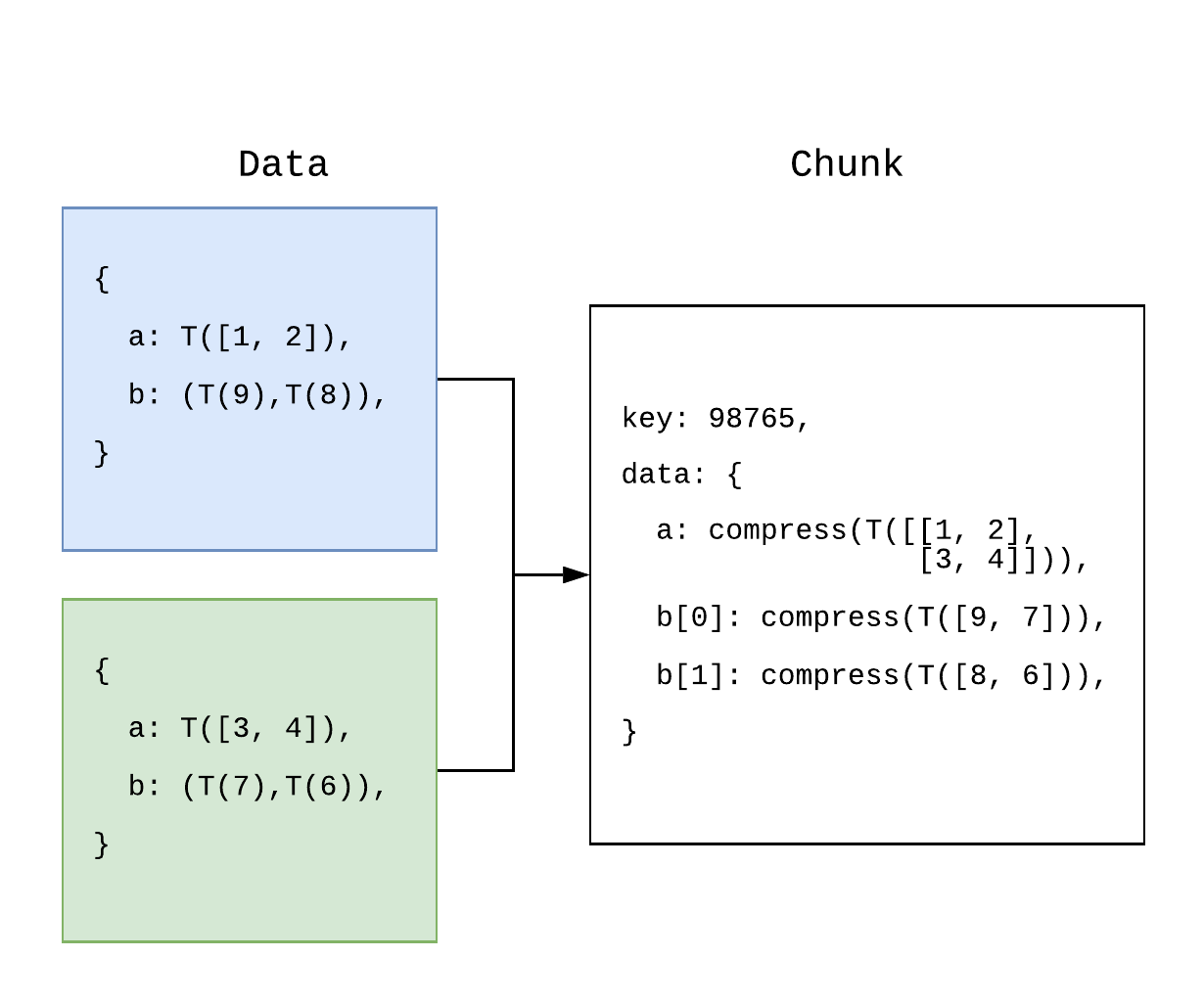}
        \centering
        \caption{
            A \Chunk{} is constructed from one or more sequential experience data objects.
            Tensors, shown as \code{T(\ldots)}, are batched column-wise and compressed.
        }
        \label{fig:chunking}
    \end{subfigure}
    \par\bigskip 
    \begin{subfigure}{\columnwidth}
        \centering
        \includegraphics[scale=0.75]{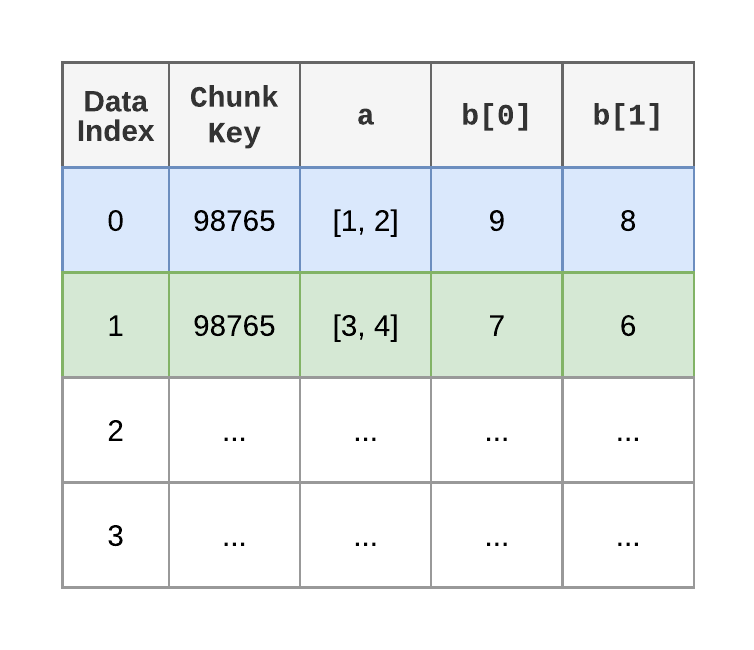}
        \caption{
            The flattened stream of data can be viewed as a two-dimensional table.
            Rows represent data elements, columns are fields in the signature, and a cell is the (Tensor) content of the field in a data element.
        }
        \label{fig:flattened_table}
    \end{subfigure}
    \caption{An example that shows how data elements are combined into a
    \Chunk{}, and a tabular layout view of this data.}
\end{figure}

\section{Design}
\label{sec:design}

\begin{figure}[t]
    \includegraphics[scale=0.4]{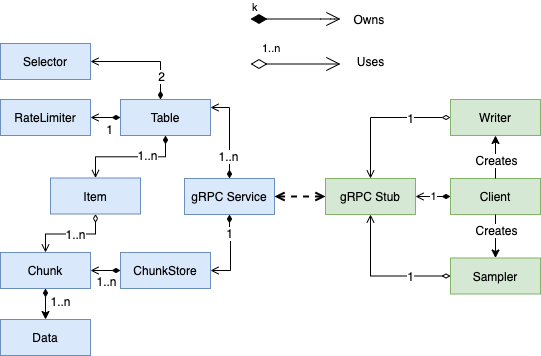}
    \centering
    \caption{High level schema of how the most important classes relate to each other.}
\label{fig:system}
\end{figure}

At its core, Reverb is a data store that exposes a gRPC service with an
API for clients to write raw tensor data. Data is written to a \ChunkStore{} as parts
of a \Chunk{}, \Chunks{} can be referenced by \Items{}, and each \Item{} is owned by a \Table{}.
The \Table{} encapsulates \Items{} and controls sample and insert requests,
using a \RateLimiter{} and two \Selectors{} (one \Sampler{} and one \Remover{}).
These Components are independently configured and can be mixed and matched to
provide a high degree of flexibility and a wide range of different behaviors.

\subsection{Chunking and the ChunkStore}
\label{sec:chunking}

\begin{figure}[t]
    \includegraphics[scale=0.38]{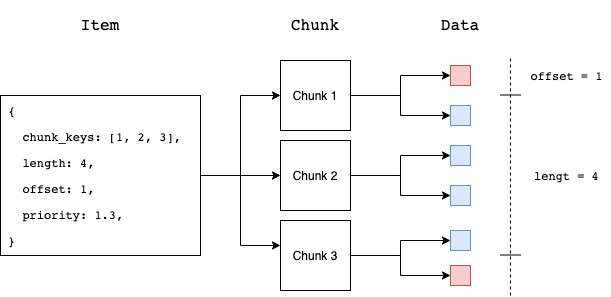}
    \centering
    \caption{
        An \Item{} references experience using one or more \Chunks{}.
        \code{Offset} and \code{length} determine the exact steps selected.
    }
    \label{fig:item}
\end{figure}

Sequential data generated by an RL environment often contain a high degree of similarity across steps.
As an example, consider the pixels in two sequential frames generated in Atari.
Many pixels will be exactly the same in both frames.
Reverb exploits this similarity by placing sequential data elements
into \Chunks{} and applying column wise concatenation and compression
(see Figure~\ref{fig:chunking} for an illustration).

Reverb expects data to be provided as a stream of sequential data elements.
Each element is a nested object whose leaf nodes are tensors.
The signature, i.e the nested object's structure, shapes and data types of its leaf tensors,
must remain the same across data elements in the stream.
This means that, when flattened, the stream of data elements can be thought of as a
two dimensional matrix where each row represents a data element and each column represents a
field in the signature (see Figure~\ref{fig:flattened_table}).

\Chunks{} reduce the memory footprint and network usage in two ways.
First, sequential data elements are grouped into a \Chunk{} and compressed.  
Second, the \Chunk{} abstraction over the raw data allows multiple
\Items{} (which can belong to different \Tables{}) to reference the
same underlying data instead of maintaining separate copies.

The \Chunks{}
are owned by a \ChunkStore{}, as seen in Figure~\ref{fig:system}. The \ChunkStore{} uses reference counting to track the number of \Items{} that reference each \Chunk{}.
When the reference count reaches zero, the \Chunk{}'s memory is automatically freed.
Decoupling data deallocation from the (mutex protected) operations on \Tables{} is important for high and stable throughput.

The \ChunkStore{} and \Table{} were designed to avoid strong coupling in their respective implementations.
Currently, \Items{} reference and retrieve data through \Chunks{} and the \ChunkStore{} stores content
in memory. The flexible design allows for alternative storage solutions to be incorporated with minimal
impact on the wider system. An example would be customizing Reverb to to access large datasets
stored on disk by modifying \Chunks{} to reference externally stored data rather than owning it. The
client would lookup the data using the references received the sampled \Chunks{}.

\subsection{Tables and Items}
A Reverb server consists of one or more \Tables{}. A \Table{} holds 
 \Items{}, defines \Selectors{} for sampling
and removal, and defines both a maximum item capacity and a \RateLimiter{}.
\Chunks{} are only sent to the server when the client signals that an \Item{} should be created.  Sampling from a \Table{} will send a number of \Items{} (along with the \Chunk{} data they reference) to the client. 
Communication overhead may occur if the number of data elements ($K$) in each \Chunk{} does not evenly divide by the number of elements in an \Item{} ($N$). 
In this case, the \Item{} only makes use of the first $N - K$ steps in the last \Chunk{}.
However, all of the $K$ steps will be sent when sampling.
This overhead can be avoided by setting $K$ and $N$ such that $N \bmod K = 0$,
as shown in Figure~\ref{fig:item}.

An \Item{} in a \Table{} is composed of:
\begin{itemize}
    \item A unique \code{key}.
    \item A \code{priority} that may be used for sampling and/or removal; clients can update this value.
    \item References to a sequence of \Chunks{}.
    \item The current number of times the \Item{} has been sampled.
\end{itemize}

An \Item{} is removed from a table in two situations: (1) when the \Item{} reaches the maximum number of times it can be sampled (which is deined when creating the \Table{}) or (2) when the \Table{} reaches its capacity limit.
In the latter case, the \Remover{} automatically selects an item to be removed before any more are inserted.

\subsection{Selectors}
A \Selector{} is a strategy used to select an item in a \Table{}.
A \Table{} uses two \Selectors{} (see Figure~\ref{fig:system}):
\begin{itemize}
    \item The \Sampler{} is used to select \Items{} when the client requests them. 
    \item The \Remover{} is used to select an \Item{} to remove when the \Table{} is full.
\end{itemize}

\Selectors{} are responsible for building and maintaining their own internal state by observing all operations on its parent \Table{}.
They must use only their internal state to decide which \Item{} to select.
Importantly, for performance reasons, they cannot make decisions based
on the content of the data that each \Item{} contains.

Reverb comes with the following \Selector{} strategies and can be extended to support others:

\textbf{FIFO:} First-in-first-out.
Provides queue-style sampling when used as \Sampler{}.
When used as \Remover{}, the oldest \Item{} is removed from the \Table{} when full.

\textbf{LIFO:} Last-in-first-out.
Selects the most recently inserted \Item{} from the \Table{}.
It is a suitable \Sampler{} for many on-policy algorithms.
As a \Remover{} it will leave the oldest \Items{}, thus allowing the \Table{} to act like a stack.

\textbf{Uniform:} Selects each \Item{} in the \Table{} with the same probability.
It is commonly used as a \Sampler{} in combination with a FIFO \Remover{}, creating a fixed size ER of the most recently generated experience.

\textbf{Max/Min Heap:} Selects the \Item{} with highest/lowest priority.
When used as \Sampler{}, this gives \Table{} priority queue behavior.
Useful as a \Remover{} in order to create a \Table{} that acts like a view of the highest priority data across longer time spans.

\textbf{Prioritized:} It implements the algorithm described in Prioritized Experience Replay \cite{schaul2015prioritized}. The probability of selecting an \Item{} with priority $p_i$ from a \Table{} of $N$ \Items{} is 
\[
    p_i = \frac{p_i^C}{ \sum_{k=1}^{N} p_k^{C}}
\]
where $p_k$ is the priority of \Item{} $k$ and $C$ is configurable constant.

\subsection{Rate Limiting}
\label{sec:rate_limiting}

The \RateLimiter{} controls when \Items{} are allowed to be inserted and/or sampled from a \Table{}.
The \RateLimiter{} monitors two aspects of the \Table{}: 1) the number of \Items{}
in the table, and 2), the \SPI{} defined as:

\[
    \SPI{} = \frac{\NumSampledItems{}}{\NumInsertedItems{}} 
\]

\RateLimiters{} define:
\begin{enumerate}
    \item The minimum number of \Items{} that must be in a \Table{} before sampling can begin.
    \item The target \SPI{} ratio
    \item The upper and lower bounds for the target \SPI{} ratio.
\end{enumerate}

These three parameters are used to define a variety of \RateLimiters{} offering a wide range of behaviors.

\begin{figure}
\centering
    \begin{subfigure}{0.495\columnwidth}
        \includegraphics[scale=0.75]{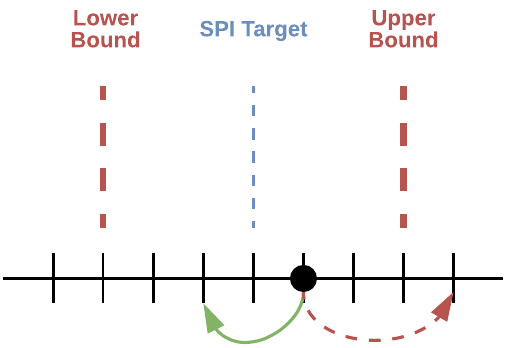}
        \caption{Inserts are blocked because they would exceed the upper SPI limit.}
        \label{fig:ratelim_1}
    \end{subfigure}
    \begin{subfigure}{0.495\columnwidth}
        \includegraphics[scale=0.75]{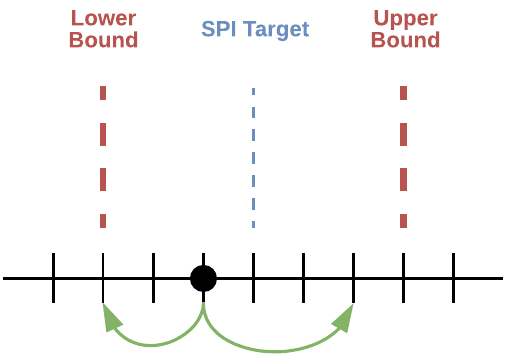}
        \caption{Once a sample takes place, decreasing \SPI{} by -2, inserts and sampling are both allowed.}
        \label{fig:ratelim_2}
    \end{subfigure}
\caption{
Illustrative of a \RateLimiter{} with $ {SPI} = {\frac{3}{2}} $.
The cursor is moved $ +3 $ by inserts and $ -2 $ by samples.}
\label{fig:ratelim}
\end{figure}

\RateLimiters{} block sampling from a \Table{} when either: 1) there are not enough \Items{} or 2) the sample would result in an \SPI{} that exceeds the upper bound.
Similarly, inserts are blocked when they would result in an \SPI{} below the lower bound. See Figure~\ref{fig:ratelim} for an illustration of how \RateLimiters{} can block inserts and samples.

The following \RateLimiters{} are available in Reverb\footnote{
The list includes pre-configured \RateLimiters{} available in the Python API.
Note that the implementation remains the same and only the
\MinSize{}, \SPI{} and bounds differ.}:

\textbf{SampleToInsertRatio}: Allows users to specify a target \SPI{} and the
minimum number of \Items{} that the \Table{} must contain before sampling is allowed. 
A single float (\code{error\_buffer}) is used to define a symmetric upper and lower bounds on the $SPI$ ratio.
Larger values avoid unnecessary blocking when the system is more or less in equilibrium.

\textbf{MinSize}: Enforces that the \Table{} must contain a certain
number of \Items{} before sampling can begin.
The \SPI{} is ignored simply by setting the upper and lower bounds
to \code{DBL\_MAX} and \code{DBL\_MIN} respectively.

\textbf{Queue}: Used to turn \Table{} into a queue-like data structure.
Inserts are allowed until the queue is full and samples are allowed unless the queue is empty.
A single parameter, \code{queue\_size}, is exposed to configure the \RateLimiter{}.
The minimum number of \Items{} is set to 0, \SPI{} is set to 1
with lower and upper bounds set to 0 and \code{queue\_size} respectively.
When combined with FIFO \Selectors{}, the \Table{} behavior becomes that of a queue. When
used in conjunction with LIFO \Selectors{}, it becomes a stack.

Finally, \RateLimiters{} control the flow of \Items{}, and not that of the data the \Items{} reference.
Adjusting the number of data elements that each \Item{} references, 
e.g. using 16 instead of 4 Atari frames changes the amount of data flowing through the system,
even though the \SPI{} remains constant.
For example, if $SPI = 2$, the \Table{} capacity is 10, and an \Item{} references 4 Atari
frames, then the \Table{} will hold 40 frames\footnote{Assuming that all items
reference disjoint sequences of data.} and a learner (consumer) will see 4 frames
for each \Item{} it samples. Changing the number of frames each \Item{}
references to 16 increases the total frames in the \Table{} to 160, and the data
consumer will now see 16 frames for each sampled \Item{}.

\subsection{Table Extensions}
\Tables{} support a \code{TableExtension} API which provides the ability to execute
extra actions as part of the atomic operations of the parent \Table{}.
All of these actions are executed while holding the \Table{} mutex, so their latency is critical.
This API can be used, for example, to write extensions that provide statistics about the amount of
data that is inserted and sampled from Reverb, and diffuse priority updates to
neighboring \Items{}~\cite{gruslys2017reactor}. 

\subsection{Sharding}

The Reverb server can be scaled horizontally by adding more independent servers.
When scaling a system to a multi-server configuration, each Reverb server remains unaware
of the others and data is neither replicated nor synchronized across servers.
Similarly, checkpointing (Section~\ref{sec:checkpointing}) is managed independently and
multi-server configurations are not more nor less robust against data loss than a single server.
However, scaling the system horizontally is trivial due to the independence between servers.

When used in combination with a gRPC compatible load balancer, client operations
(Section \ref{sec:client}) are distributed across the servers in a round robin fashion. 
When sampling, each client manages pool of server connections.
Samples are requested from multiple servers in parallel and the results are merged
into a single stream of sampled \Items{}. This mitigates the effects of long-tail latency
and creates fault tolerance against individual server failures.

Since Reverb servers are independent, each can be configured differently, e.g. with different rate limiters.
A separate client can then be created for each server (rather than pooling the results) allowing for maximal control.

\subsection{Checkpointing}
\label{sec:checkpointing}

The state and content of both the \ChunkStore{} and \Tables{} can be serialized and stored to disk as a 
checkpoint. Potential data loss in the event of unexpected server failures can be limited through the use of periodic checkpointing. Stored checkpoints can be loaded by Reverb servers at construction time.

In order to allow Reverb checkpoints to be aligned with the wider systems, e.g., network weights,
the creation of checkpoints is triggered through a gRPC call from a Reverb client.
During the checkpointing process, the server blocks all incoming insert, sample, update, and delete requests.
This process could potentially last for multiple minutes and is dependent on the amount of
data being written and disk performance.

\subsection{Reverb Client}
\label{sec:client}
The Reverb Client wraps the gRPC client to provide a higher level API for writing, modifying,
and reading data from the server.

A \Sampler{} manages a pool of long lived gRPC streams. 
Each stream fetches samples from a single \Table{} at a flow controlled rate that can be adjusted using the \code{max\_in\_flight\_samples\_per\_worker} parameter.
Setting this parameter to 1 means that the next sample is not requested until the previous one has been consumed. 
Setting the parameter higher gives the \Sampler{} more flexibility to prefetch samples, which generally leads to higher throughput. 
For more details about how to retrieve data efficiently, see Section (\ref{sec:tfdata}).

A \Writer{} is used to stream sequential data to the server and insert \Items{} into one or more \Tables{}.
A single gRPC stream is used throughout the lifetime of a \Writer{} object.
Data is written using the \code{append} method and \Items{} are created using \code{create\_item}.
When \code{append} is called, the data is pushed to a local buffer.
Once the buffer is full, a \Chunk{} is constructed (See Section~\ref{sec:chunking}).
The \Chunk{} is then transmitted to the server over the open gRPC stream.
Similarly, \Items{} created with \code{create\_item} are pushed to a local buffer until all its
referenced \Chunks{} have been transmitted to the server. Waiting for the \Chunk{} to be sent
before \Items{} makes it safe for multiple items to reference the same data without sending it more than once.

\subsection{Efficient Sampling with the \TFData{} API}
\label{sec:tfdata}
Reverb utilizes the \TFData{} framework \cite{tensorflow2015-whitepaper} to provide pipelined high
throughput data sampling directly to training modules. 

\TFData{} is a proven high performance and scalable input pipeline solution for transforming
and feeding data. It is used in TensorFlow 2 as part of stand-alone and distributed training
configurations. Beyond TensorFlow, it has also been commonly used in 
JAX~\cite{frostig2018compiling} as part of training workflows.

Reverb implements the \ReverbDataset{}: a fast, C++-based mechanism for reading and postprocessing data from Reverb servers.
Iterators created by \ReverbDataset{} each wrap a \Sampler{}.
Each \Sampler{} in turn maintains a configurable number of long-lived streams.
If a single stream is used, the dataset feeds data in exact order from a Reverb server.
This is necessary when the associated \Table{}'s, \Sampler{}, and \Remover{} are configured to use deterministic \Selectors{} such as FIFO.

One idiosyncrasy to be aware of is that the system can appear to deadlock if the underlying \Table{}'s
size goes below the minimum size or the sample-to-insert ratio grows too large. This is
not a bug. Reverb is designed to enforce these parameters. The situation can occur
organically if the training processes are consuming \Items{} at a ratio the actors
cannot sustain. The situation may resolve itself quickly as new data is inserted or
be permanent due to a downstream failure. The perceived deadlock can be managed by setting the
$\code{rate\_limiter\_timeout\_ms} \ge{} 0$ when creating the dataset. 
If a sample is requested but not enough data is received within the time limit, the reverb
service will signal to the iterator that it is safe to end the sequence. This is
similar to "reaching the end of the file" on Datasets that read from on-disk storage. 

\section{Examples}
\label{sec:examples}
This section contains basic examples of core Reverb's features. See Appendix~\ref{appendix:examples}
for examples of more specific RL use cases.

\subsection{Overlapping trajectories}
The following example shows how to write trajectories of length 3 that overlap by 2 timesteps. 

\begin{minipage}[]{\linewidth}
\begin{minted}[fontsize=\footnotesize]{python}
NUM_TIMESTEPS = 3
with client.writer(NUM_TIMESTEPS) as w:
 ts, a, done = env_step(None)
 step = 0
 while not done:
   writer.append((ts, a))
   if step >= 2:
     # Items reference the 3 most
     # recently appended timesteps
     # and have a priority of 1.5.
     writer.create_item(
         table='my_table_a',
         num_timesteps=NUM_TIMESTEPS,
         priority=1.5)
   ts, a, done = env_step(action)
   step += 1
\end{minted}
\end{minipage}

\subsection{Multiple priority tables}
The example below utilizes two different tables to create trajectories of different lengths.

\begin{minipage}[]{\linewidth}
\begin{minted}[fontsize=\footnotesize]{python}
with client.writer(3) as writer:
 ts, a, done = env_step(None)
 step = 0
 while not done:
   writer.append((timestep, action))
   if step >= 1:
     writer.create_item(
         table='my_table_a',
         num_timesteps=2,
         priority=1.5)
   if step >= 2:
     writer.create_item(
         table='my_table_b',
         num_timesteps=3,
         priority=1.5)
    ts, a, done = env_step(action)
   step += 1
\end{minted}
\end{minipage}

\section{Performance}
\label{sec:performance}

\begin{figure*}[t]
    \centering
    \begin{subfigure}{\columnwidth}
        \centering
        \includegraphics[scale=0.17]{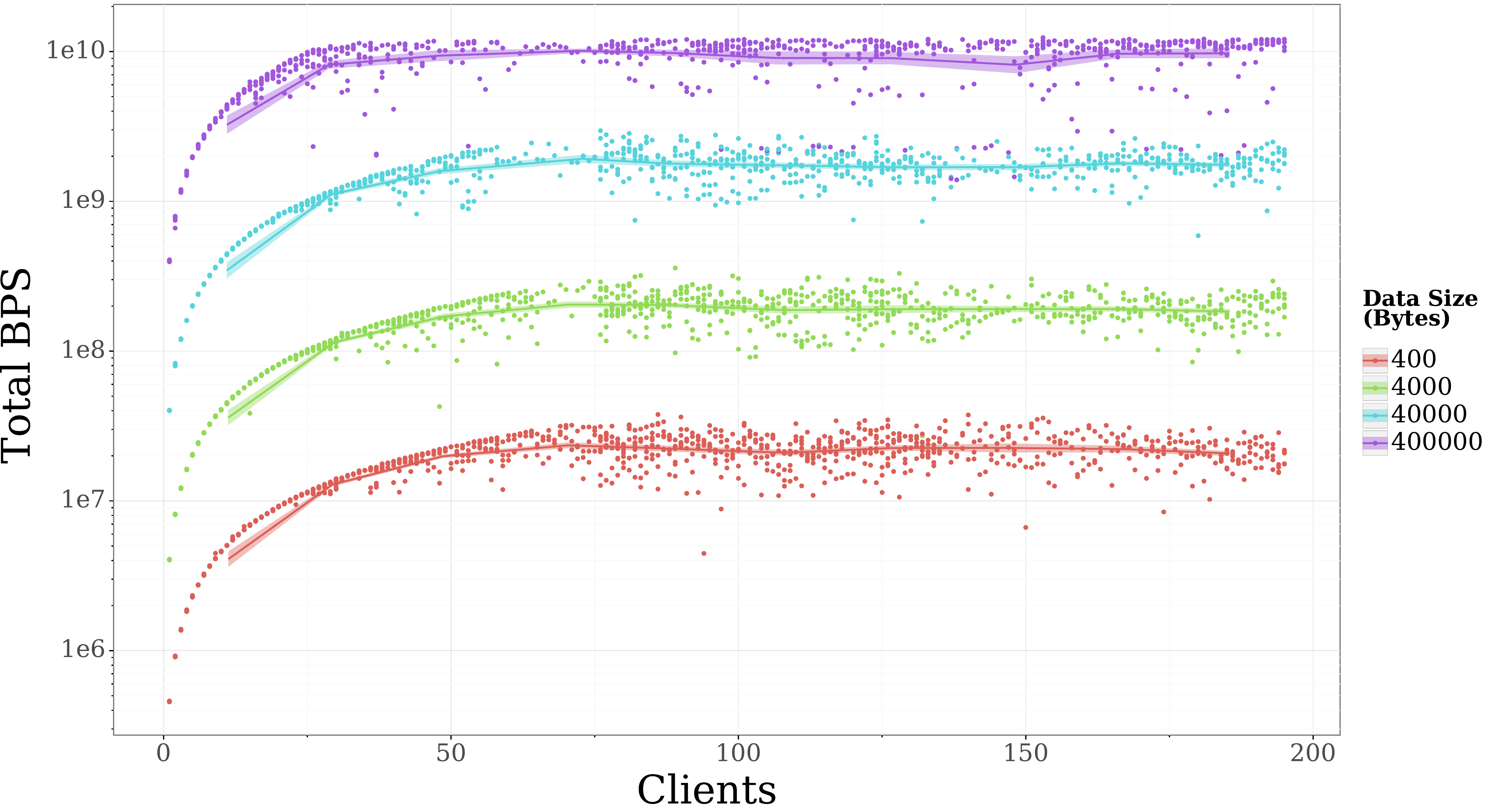}
        \captionsetup{width=.8\columnwidth}
        \caption{
            Bytes Per Second (BPS) that the server handles plotted against the number of connected clients.
        }
        \label{fig:inserting_bps}
    \end{subfigure}
    \begin{subfigure}{\columnwidth}
        \centering
        \includegraphics[scale=0.17]{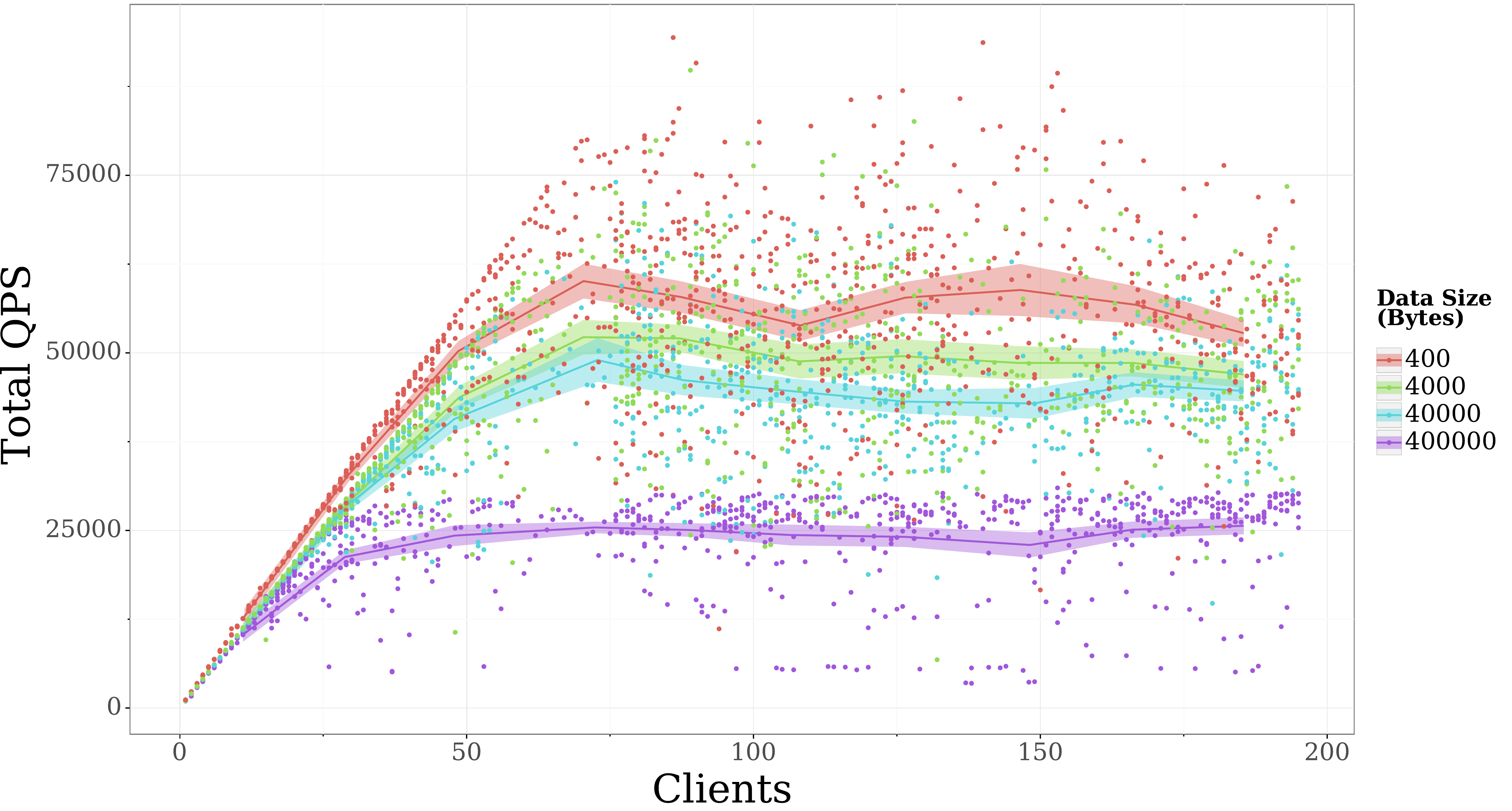}
        \captionsetup{width=.8\columnwidth}
        \caption{
            Queries Per Second (QPS) that the server handles plotted against the number of connected clients.
        }
        \label{fig:inserting_qps}
    \end{subfigure}
    \caption{
        Server performance benchmarks for inserting data.
        Both \ref{fig:inserting_bps} and \ref{fig:inserting_qps} plots the results for tensor payload sizes between 400 Bytes and 400 kB.}
\label{fig:inserting_stats}
\end{figure*}

\begin{figure*}[t]
    \centering
    \begin{subfigure}[b]{\columnwidth}
        \centering
        \includegraphics[scale=0.17]{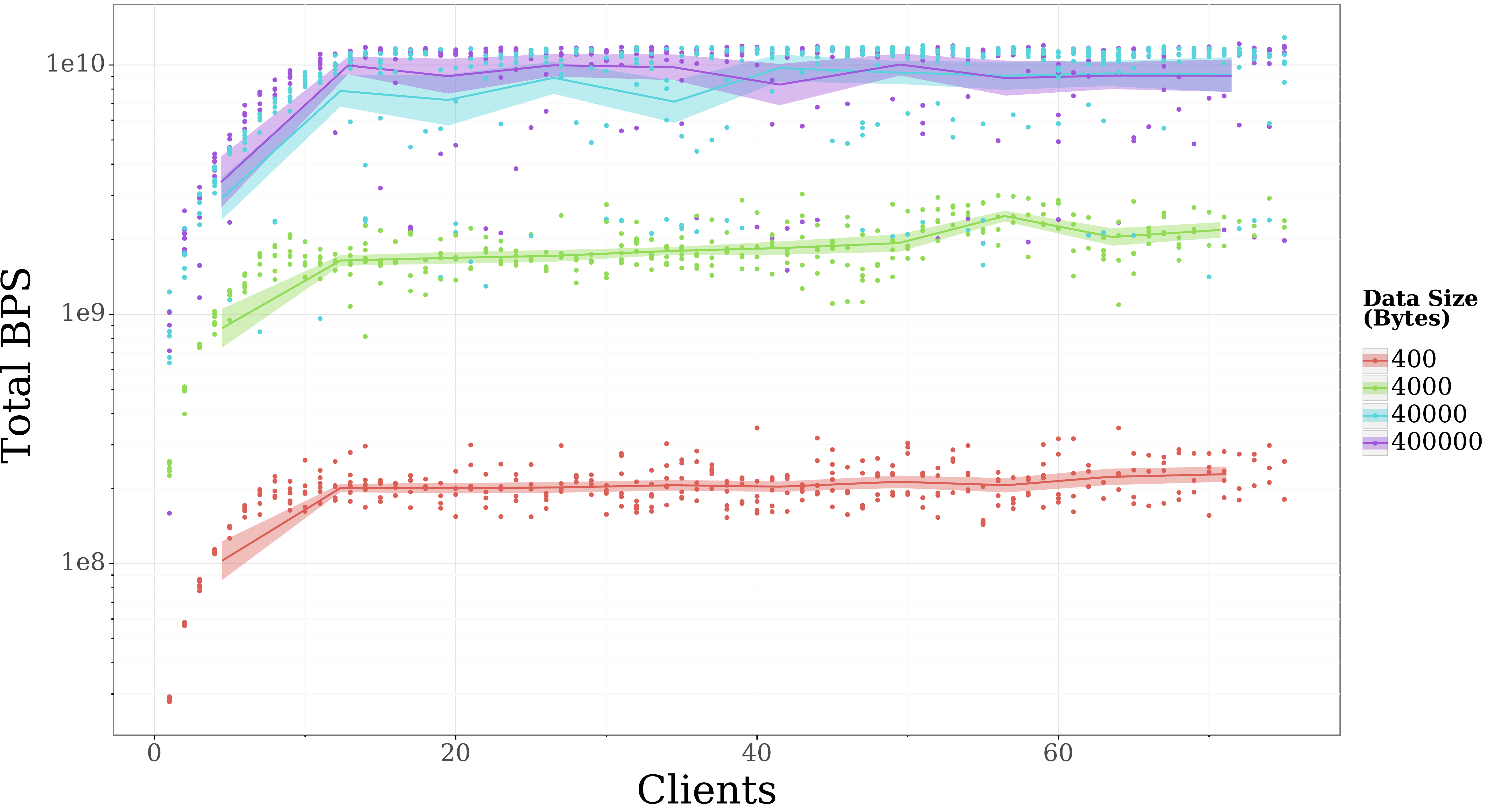} 
        \captionsetup{width=.8\columnwidth}
        \caption{
            Bytes Per Second (BPS) that the server handles plotted against the number of connected clients.
        }
    \end{subfigure}
    \begin{subfigure}[b]{\columnwidth}
        \centering
        \includegraphics[scale=0.17]{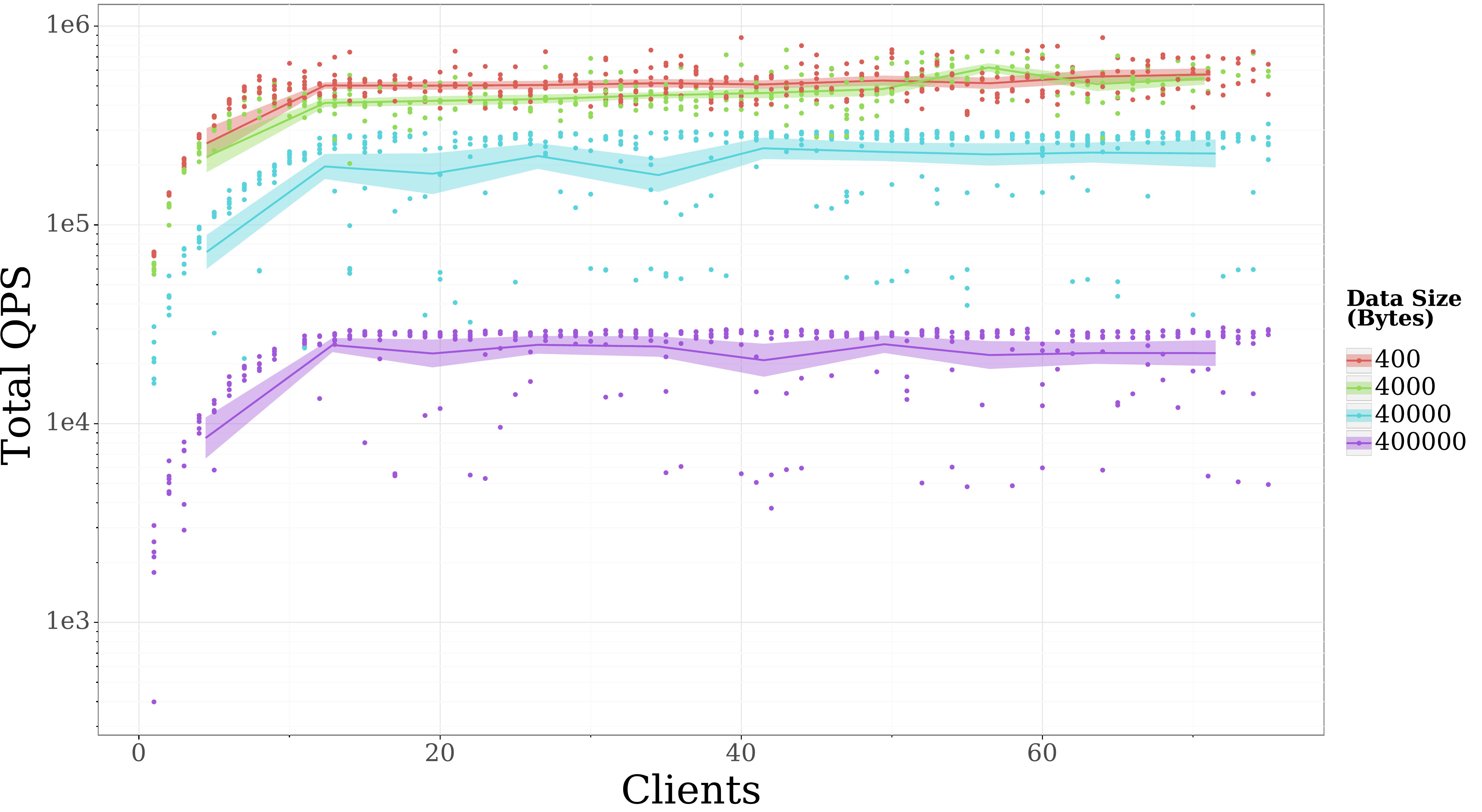} 
        \captionsetup{width=.8\columnwidth}
        \caption{
            Queries Per Second (QPS) that the server handles plotted against the number of connected clients.
        }
    \end{subfigure}
    \caption{Server performance benchmarks for sampling data.
    Each plot in~\ref{fig:inserting_bps} and~\ref{fig:inserting_qps}
    is for a different tensor payload size, from 400 Bytes to 400 kB.}
\label{fig:sampling_stats}
\end{figure*}

This section presents two sets of benchmarks designed to explore the scaling characteristics of a
single Reverb server. We want to emphasize that the raw numbers reported are specific to this particular
benchmark setup and carry little significance by themselves.
We encourage the reader to focus their attention on the patterns that occur across the benchmarks and the
generic conclusions that can be drawn from these observations.
 
The benchmarks have been explicitly designed to overload the server under the most unfavourable conditions.

\begin{enumerate}
    \item {
        Each data element is a single \code{float32} tensor whose values have been randomly sampled
        from a uniform distribution over the half-open interval $[0, 1)$.
        Random tensors were intentionally used to negate variability introduced by compression.
        Reported Bytes/s are much lower than what would be observed in the exact
        same setup modulo the source of data. For example, in Atari we observe compression rates of
        up ~90\% in sequences of 40 frames. The effective throughput would therefore be up to 
        10x higher in that scenario compared with these benchmarks utilizing random synthetic data.
    }
    \item {
        Chunk and sequence length is 1 resulting in \Items{} not sharing data.
    }
    \item {
        Clients solely generate load as fast as possible.
    }
\end{enumerate}

The benchmarks are distributed with each client running on a different machine.
All machines are located in the same data center and communicate with the server over network shared by other citizens of the data center.
We increase the number of clients until the combined load far exceeds the server's capabilities.
The same test was performed with data payloads across four order of magnitudes: 400B, 4kB, 40Kb, and 400kB.

Each benchmark tests the upper bound of a single Reverb server's efficiency with all clients either
inserting (Section~\ref{sec:inserting}) or sampling (Section~\ref{sec:sampling}). The results are reported
in Total Bytes Per Second (BPS) and Queries (i.e \Items{}) Per Second (QPS) that the single Reverb server
could process given the number of clients.

\subsection{Inserting}
\label{sec:inserting}

For the inserting benchmark, all clients write data to the same server. The benchmark was run
with 1 to 200 clients. Examining Figure~\ref{fig:inserting_stats} leads us to the following
general conclusions regarding the scalability of Reverb:

\begin{enumerate}
    \item The server can handle approximately either 11 GB/s or 60k Items/s.
    \item The throughput scales perfectly (i.e linearly) with the number of clients until either the BPS or QPS limit is reached.
    \item Overloading the server with additional clients does not degrade performance.
\end{enumerate}

The linear scaling behavior is the important result of the benchmark. The Figure~\ref{fig:inserting_stats}
illustrates that the number of concurrent clients have no negative impact on each other unless the
combined QPS or BPS load exceeds the server limits. Once the limit is reached, the server will distribute
its available resources across the clients. In this benchmark, the server continued to do this effectively
for the max number of clients (200) tested.

In the real world clients are unlikely to be writing data as quickly as the synthetic benchmark due to 
needing to generate data in an environment, e.g. Atari. Consider for example a setup were each
client is only able to generate data at 1/4 the speed of the benchmark clients. In this same compute
environment, we would expect linear scaling to continue well beyond 200 clients until the combined
QPS or BPS load exceeds the server limits. The ability to scale to large number of clients without
degrading the performance is important for scenarios were the data generation is more expensive or slower.

\subsection{Sampling}
\label{sec:sampling}
For the sampling benchmark, each client samples data as quickly as possible from a single Reverb server. The
benchmark was run with 1 to 200 clients. From Figure~\ref{fig:sampling_stats} it is apparent that:

\begin{enumerate}
    \item The server can handle approximately either 11 GB/s or 600k Items/s.
    \item The throughput scales almost perfectly (i.e linearly) with the number of clients
    until either the BPS or QPS limit is reached.
    \item Overloading the server with additional clients does not degrade performance.
\end{enumerate}

We observe almost identical scaling characteristics as seen in the inserting benchmark
(Section~\ref{sec:inserting}) but with a tenfold increase in maximum QPS capacity.
The difference in QPS capacity can most likely be attributed to optimizations intended
to reduce \Table{} mutex contention, which, at the time of writing this paper, is only implemented for sampling
and not insertions. 

\subsection{Summary}
\label{sec:summary}

The maximum BPS is 11 GB/s for both insertion and sampling. This limit is observed at multiple
payloads sizes and strongly indicates that the observed BPS limitations cannot be attributed to Reverb.
The limitation is most likely the result of network bandwidth constraints.

To test our hypothesis that QPS for inserts is currently limited by mutex contention, we setup a
sharding experiment detailed in Appendix~\ref{appendix:sharding}. The load was sharded across 
multiple \Tables{} on the same server. This improved the maximum insert QPS by approximately 200\%,
and convinced us that the insert QPS limitations are due to mutex contention.

\section{Conclusion}
\label{sec:conclusion}

Reverb is an efficient and easy-to-use data storage and transport system designed for machine learning research.
 Reverb's flexible API allows a single server to scale effortlessly from one to hundreds of concurrent clients under unfavorable conditions while handling loads of $O(100k)$ QPS and at least 11 GB/s of compressed data.
This flexibility, paired with the scaling characteristics, enables researchers to run experiments using a single-process or thousands of machines with the same Reverb setup.
As a result, Reverb is currently being used by hundreds of researchers for a wide variety of RL experiments.

\bibliography{main.bib}
\bibliographystyle{mlsys2020}

\clearpage
\appendix
\section{Reinforcement Learning Examples}
\label{appendix:examples}

\subsection{Acme: D4PG}
The D4PG agent in Acme\footnote{\url{https://git.io/Jtuw5}} uses a Reverb \Table{} with an Uniform \Sampler{} and a FIFO \Remover{}.
Items are thus selected uniformly from the table, and the oldest \Item{} is removed when the buffer reaches \code{MAX\_REPLAY\_SIZE}.

\begin{minipage}[]{\linewidth}
\begin{minted}[fontsize=\footnotesize]{python}
table = reverb.Table(
    name=TABLE_NAME,
    sampler=reverb.selectors.Uniform(),
    remover=reverb.selectors.Fifo(),
    max_size=MAX_REPLAY_SIZE,
    rate_limiter=\
        reverb.rate_limiters.MinSize(1))
server = reverb.Server([table])
\end{minted}
\end{minipage}

The \code{MinSize(1)} rate limiter makes the sampler to block until there is at least one element in the table.
It uses the default value for \code{max\_times\_sampled} (which is 0), so each item can be sampled any number of times until it is removed by the FIFO remover.
Each item is a n-step transition which is defined in Acme as a transition that accumulates the reward and the discount for n steps.

\subsection{TF-Agents: Distributed SAC}
The Distributed SAC implementation in TF-Agents\footnote{\url{https://git.io/JtuwH}} uses two Reverb \Tables{}.

The first one is a Variable Container.
This \Table{} holds the model parameters most recently exported by the learner.
By sampling from this table, the actors are able to update the model parameters, which they use to run the inference.
Only the most recently exported parameters are used and thus the \Table{} have a maximum size of 1 and a FIFO \Remover{} is used.
Since the \Table{} only contains at most 1 \Item{}, any \Selector{} will work as a \Sampler{}.
To allow the same parameters to be sampled as many times as needed, (\code{max\_time\_sampled} is set to 0, i.e no limit.
The \code{MinSize(1)} ensures that actors block until the learner have exported the first version of the parameters.

\begin{minipage}[]{\linewidth}
\begin{minted}[fontsize=\footnotesize]{python}
# Stores policy parameters.
reverb.Table( 
    name='VARIABLE_CONTAINER',
    sampler=reverb.selectors.Uniform(),
    remover=reverb.selectors.Fifo(),
    rate_limiter=\
        reverb.rate_limiters.MinSize(1),
    max_size=1,
    max_times_sampled=0,
)
\end{minted}
\end{minipage}

The Experience Replay has a similar configuration to D4PG.
By default, it also uses a \code{MinSize} \RateLimiter{}, which will stop the sampling until there are enough elements in the \Table{}. 
However, since each item can be sampled any number of times, it will only block the sampler at the beginning.
The rate limiter here can also be configured as a \code{SampleToInsertRatio}, which sets both the minimum size to start sampling, and the allowed number of samples per insert, which enables more fine-grained flow control.

\begin{minipage}[]{\linewidth}
\begin{minted}[fontsize=\footnotesize]{python}
samples_per_insert_tolerance = \
    _TOLERANCE_RATIO * samples_per_insert
error_buffer = (
    min_size *
    samples_per_insert_tolerance
)
experience_rate_limiter = 
    reverb.rate_limiters.SampleToInsertRatio(
        min_size_to_sample=min_size,
        samples_per_insert=samples_per_insert,
        error_buffer=error_buffer
)

reverb.Table(
  name=reverb_replay_buffer.DEFAULT_TABLE,
  sampler=reverb.selectors.Uniform(),
  remover=reverb.selectors.Fifo(),
  rate_limiter=experience_rate_limiter,
  max_size=FLAGS.replay_buffer_capacity,
  max_times_sampled=0,
  signature=replay_buffer_signature,
)
\end{minted}
\end{minipage}

\section{Benchmark with multiple \Tables{}}
\label{appendix:sharding}

\begin{figure}
    \centering
    \includegraphics[scale=0.17]{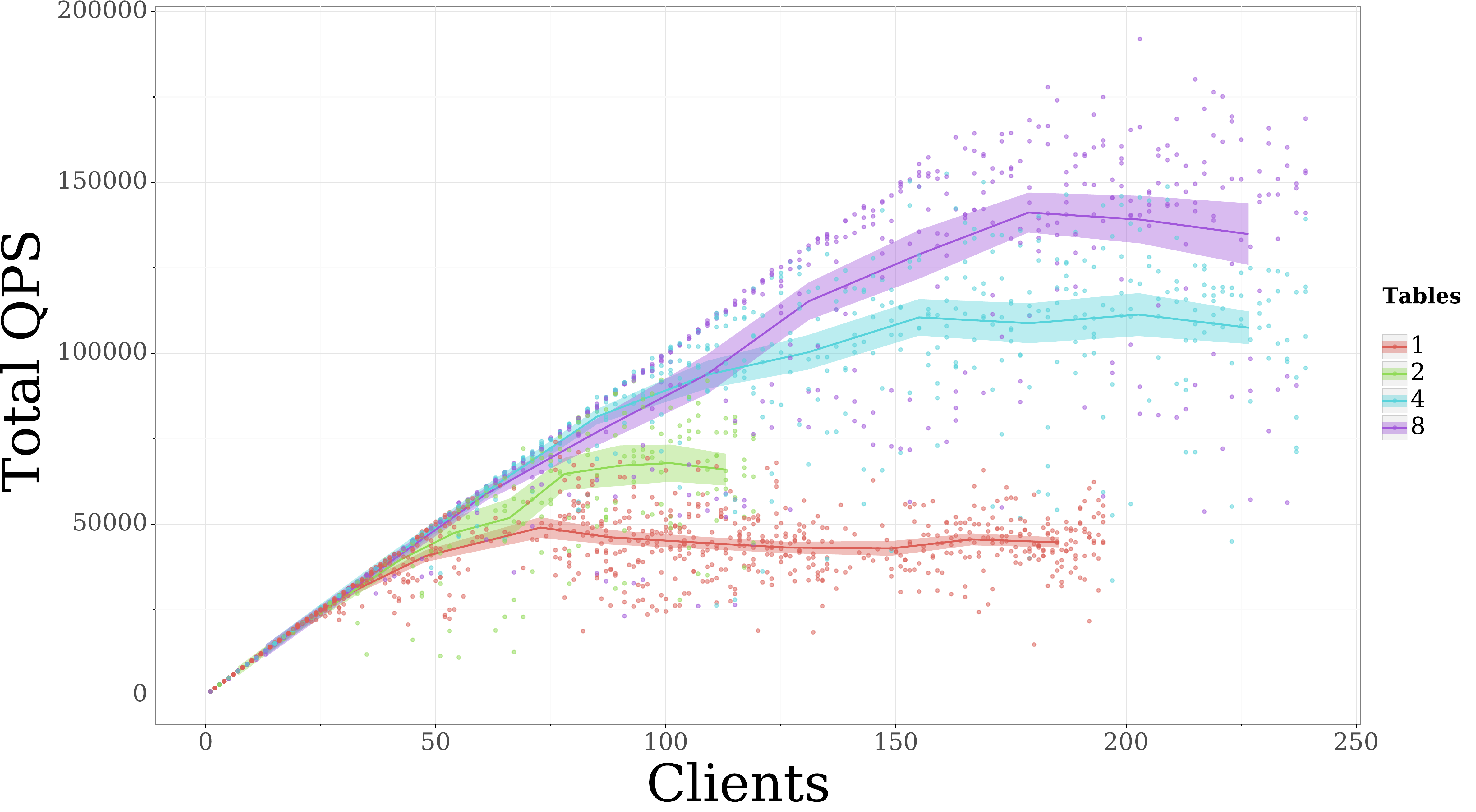} 
    \caption{
        Items inserted per second (QPS) that a single server handles plotted against the number of connected clients.
        The maximum QPS is increased by approximately 200\% when splitting the \Table{} into 8 shards. 
    }
    \label{fig:sharded_qps}
\end{figure}

This section tests the hypothesis that the difference in QPS capacity between the insert and sample benchmarks can be explained by optimizations of the \Table{} mutex usage (which has been implemented only for sample operations).
If mutex contention explains the difference then spreading the load over multiple \Tables{} should improve QPS limits even if the number of servers remains one.
We modify the benchmark described in Section~\ref{sec:inserting} by varying the number of \Tables{} held by the server and modifying the clients to round robin between the \Tables{} with each \code{create\_item}.

The benchmark is executed using 2, 4 and 8 \Tables{}.
In Figure~\ref{fig:sharded_qps}, we plot the new measurements together with the original results from Section~\ref{sec:inserting}.
We observe a three-fold improvement between 1 and 8 \Tables{}.
This strongly supports the validity of the hypothesis.

\end{document}